\documentclass{article}
\usepackage{ijcai18}

\usepackage{times}
\usepackage{xcolor}
\usepackage{soul}
\usepackage[utf8]{inputenc}
\usepackage[small]{caption}

\usepackage{graphicx}
\usepackage{subcaption}
\usepackage{url}

\title{Visualizations for an Explainable Planning Agent}

\author{
Tathagata Chakraborti$^1$ \and Kshitij P. Fadnis$^2$ \and Kartik Talamadupula$^2$ \and Mishal Dholakia$^2$\\[0.5ex]
{\Large \bf Biplav Srivastava$^2$ \and Jeffrey O. Kephart$^2$ \and Rachel K. E. Bellamy$^2$}\\[1ex]
$^1$
Computer Science Department, Arizona State University, Tempe, AZ 85281 USA\\[0.5ex]
{\tt tchakra2 @ asu.edu}\\[1ex]
$^2$IBM T. J. Watson Research Center, Yorktown Heights, NY 10598 USA\\[0.5ex]
{\tt \{ kpfadnis, krtalamad, mdholak, biplavs, kephart, rachel  \} @ us.ibm.com}\\
}
\newcommand{\fresco}{\texttt{Fresco}}
\newcommand{\jones}{\texttt{Mr.Jones}}

\newcommand{\cel}{\texttt{CEL}}
\newcommand{\ma}{M\&A}

\begin{document}

\maketitle

\begin{abstract}
In this paper, we report on the visualization capabilities of an Explainable AI Planning (XAIP) agent that can support human in the loop decision making. Imposing transparency and explainability requirements on such agents is especially important in order to establish trust and common ground with the end-to-end automated planning system. Visualizing the agent's internal decision making processes is a crucial step towards achieving this. This may include externalizing the ``brain'' of the agent -- starting from its sensory inputs, to progressively higher order decisions made by it in order to drive its planning components. 
We also show how the planner can bootstrap on the latest techniques in explainable planning to cast plan visualization as a plan explanation problem, and thus provide concise model based visualization of its plans. 
We demonstrate these functionalities in the context of the automated planning components of a smart assistant in an instrumented meeting space.
\end{abstract}

\begin{figure*}
\includegraphics[width=\textwidth]{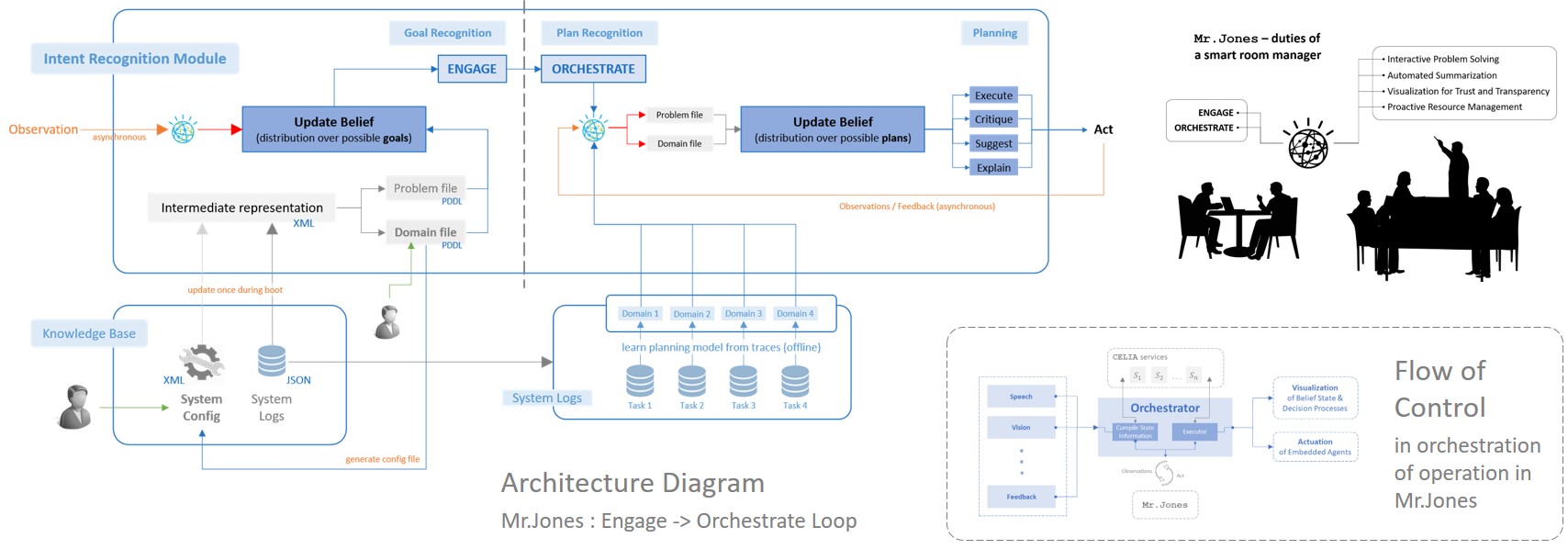}
\caption{
Architecture diagram illustrating the building blocks of \jones~--
the two main components {\em Engage} and {\em Orchestrate} situates the agent proactively in a decision support setting with human decision makers in the loop. 
The top right inset shows the different roles of \jones~as a smart room orchestrator and meeting facilitator.
The bottom right inset illustrates the flow of control in \jones~-- each service runs in parallel and asynchronously to maintain anytime response of all the individual components.
}
\label{fig:fig0}
\end{figure*}

%

\section{Introduction}



\noindent Advancements in the fields of speech, language, and search have led to ubiquitous personalized assistants like the Amazon Echo, Google Home, Apple Siri, etc. 
Even though these assistants have mastered a narrow category of interaction in specific domains, they mostly operate in passive mode -- i.e. they merely respond via a set of predefined scripts, most of which are written to specification. In order to evolve towards truly smart assistants, the need for (pro)active collaboration and decision support capabilities is paramount.
%
%
%
%
%
%
%
%
%
%
Automated planning offer a promising alternative to this drudgery of repetitive and scripted interaction. The use of planners allows automated assistants to be imbued with the complementary capabilities of being nimble and proactive on the one hand, while still allowing specific knowledge to be coded in the form of domain models. Additionally, planning algorithms have long excelled \cite{myers1996advisable,radar} in the presence of humans in the loop for complex collaborative decision making tasks.
%
%
%
%
%
%
%
%
%
%
%
%
%
%
%


\vspace{-10pt}
\paragraph{eXplainable AI Planning (XAIP)}
While planners have always adapted to accept various kinds of inputs from humans, only recently has there been a concerted effort on the other side of the problem: making the outputs of the planning process more palatable to human decision makers. The paradigm of eXplainable AI Planning (XAIP)~\cite{danmaga} has become a central theme around which much of this research has coalesced. In this paradigm, emphasis is laid on the qualities of {\em trust}, {\em interaction}, and {\em transparency} that an AI system is endowed with. The key contributions to explainability are the resolution of critical exploratory questions -- why did the system do something a particular way, why did it not do some other thing, why was its decision optimal, and why the evolving world may force the system to replan. 
%
%
%
%
%
%
%
%


%
%
%
%


\vspace{-10pt}
\paragraph{Role of Visualization in XAIP}
One of the keys towards achieving an XAIP agent is visualization. 
The planning community has recently made a concerted effort to support the visualization of key components of the end-to-end planning process: from the modeling of domains~\cite{bryce2017insitu}; to assisting with plan management~\cite{izygon2008procedure}; and beyond~\cite{radar,J.}.
For an end-to-end planning system, this becomes even more challenging since the systems state is determined by information at different levels of abstraction which are being coalesced in the course of decision making.
A recent workshop \cite{DBLP:workshop/uisp/2017} outlines these challenges in a call to arms to the community on the topic of visualization and XAIP.

%
%
%
%



\vspace{-10pt}
\paragraph{Contribution}

It is in this spirit that we present 
a set of visualization capabilities for an XAIP agent that assists with human in the loop decision making tasks: specifically in the case of this paper, assistance in an instrumented meeting space. We introduce the end-to-end planning agent, \jones, and the visualizations that we endow it with. We then provide fielded demonstrations of the visualizations, and describe the details that lie under the hood of these capabilities.

\subsection{Introducing \jones}

First, we introduce \jones, situated in the \cel ~-- the Cognitive Environments Laboratory -- at IBM's T.J. Watson Research Center. 
%
%
\jones\ is designed to embody the key properties of a proactive assistant while fulfilling the properties desired of an XAIP agent.
%
%

\subsection{\jones: An end-to-end planning system}

We divide the responsibilities of \jones~into two processes – {\bf \em Engage}, where plan recognition techniques are used to identify the task in progress; and {\bf \em Orchestrate}, which involves active participation in the decision-making process via real-time plan generation, visualization, and monitoring. 

\vspace{-10pt}
\paragraph{ENGAGE} This consists of \jones\ monitoring various inputs from the world in order to situate itself in the context of the group interaction. First, the assistant gathers various inputs like speech transcripts, live images, and the positions of people within a meeting space; these inputs are fed into a higher level symbolic reasoning component. Using this, the assistant can (1) {\em requisition} resources and services that may be required to support the most likely tasks based on its recognition; (2) {\em visualize} the decision process -- this can depict both the agent's own internal recognition algorithm, and an external, task-dependent process; and (3) {\em summarize} the group decision-making process.

\vspace{-10pt}
\paragraph{ORCHESTRATE}
This process is the decision support assistant's contribution to the group's collaboration.
This can be done using standard planning techniques, and can fall under the aegis of one of four actions as shown in Figure~\ref{fig:fig0}. These actions, some of which are discussed in more detail in~\cite{radar}, are: (1) {\em execute}, where the assistant performs an action or a series of actions related to the task at hand; (2) {\em critique}, where the assistant offers recommendations on the actions currently in the collaborative decision sequence; (3) {\em suggest}, where the assistant suggests new decisions and actions that can be discussed collaboratively; and (4) {\em explain}, where the assistant explains its rationale for adding or suggesting a particular decision. The {\em Orchestrate} process thus provides the ``support'' part of the decision support assistant. 
The {\em Engage} and {\em Orchestrate} processes can be seen as somewhat parallel to the {\em interpretation} and {\em steering} processes defined in the crowdsourcing scenarios of \cite{talamadupula2013herding,manikondaherding}. The difference in these new scenarios is that the humans are the final decision makers, with the assistant merely supporting the decision making.

\subsubsection{Architecture Design \& Key Components}

The central component -- the Orchestrator\footnote{Not to be confused with the term {\em Orchestrate} from the previous section, used to describe the phase of active participation.} -- regulates the flow of information and control flow across the modules that manage the various functionalities of the \cel; this is shown in Figure~\ref{fig:fig0}.
These modules are mostly asynchronous in nature and may be: (1) services\footnote{Built on top of the Watson Conversation and Visual Recognition services on IBM Cloud and other IBM internal services.} processing sensory information from various input devices across different modalities like audio (microphone arrays), video (PTZ cameras / Kinect), motion sensors (Myo / Vive) and so on; (2) services handling the different services of \cel; and (3) services that attach to the \jones~module. The Orchestrator is responsible for keeping track of the current state of the system as well as coordinating actuation either in the belief/knowledge space, or in the actual physical space.
%
%
%



\vspace{-10pt}
\paragraph{Knowledge Acquisition / Learning}
The knowledge contained in the system comes from two sources -- (1) the developers and/or users of the service; and (2) the system's own memory; as illustrated in Figure~\ref{fig:fig0}.
One significant barrier towards the adoption of higher level reasoning capabilities into such systems has been the lack of familiarity of developers and end users with the inner working of these technologies. With this in mind we provide an XML-based modeling interface -- i.e. a {\em ``system config''} -- where users can easily configure new environments. This information in turn enables automatic generation of the files that are internally required by the reasoning engines. Thus system specific information is bootstrapped into the service specifications written by expert developers, and this composite knowledge can be seamlessly transferred across task domains and physical configurations. 
%
%
%
%
%
%
%
%
%
The granularity of the information encoded in the models depends on the task at hand -- for example, during the {\em Engage} phase, the system uses much higher level information (e.g. identities of agents in the room, their locations, speech intents, etc.) than during the {\em Orchestrate} phase, where more detailed knowledge is needed.
This enables the system to reason at different levels of abstraction independently, thus significantly improving the scalability as well as robustness of the recognition engine.

\vspace{-10pt}
\paragraph{Plan Recognition}

The system employs the probabilistic goal / plan recognition algorithm from~\cite{ramirez2010probabilistic} to compute its beliefs over possible tasks. 
The algorithm casts the plan recognition problem as a planning problem by compiling away observations to the form of actions in a new planning problem. The solution to this new problem enforces the execution of these observation-actions in the observed order. 
This explainsmini the reasoning process behind the belief distribution in terms of the possible plans that the agent envisioned (as seen in Figure~\ref{fig:snap}).

\vspace{-10pt}
\paragraph{Plan Generation}

The \texttt{FAST-DOWNWARD} planner \cite{helmert2006fast} provides a suite of solutions to the forward planning problem. 
The planner is also required internally by the Recognition Module when using the compilation from~\cite{ramirez2010probabilistic}, or in general to drive some of the orchestration processes. 
The planner reuses the compilation from the Recognition Module to compute plans that preserve the current (observed) context.

\subsection{Visualizations in \jones}

The \cel\ is a smart environment, equipped with various sensors and actuators to facilitate group decision making. 
Automated planning techniques, as explained above, are the core component of the decision support capabilities in this setting. 
However, the ability to plan is rendered insufficient if the agent cannot communicate that information effectively to the humans in the loop. 
Dialog as a means of interfacing with the human decision makers often becomes clumsy due to the difficulty of representing information in natural language, and/or the time taken to communicate. 
Instead, we aim to build visual mediums of communication between the planner and the humans for the following key purposes --

\begin{itemize}
\item[-] {\em Trust \& Transparency -} Externalizing the various pathways involved in the decision support process is essential to establish trust between the humans and the machine, as well as to increase situational awareness of the agents. It allows the humans to be cognizant of the internal state of the assistant, and to infer decision rationale, thereby reducing their cognitive burden.
\item[-] {\em Summarization of Minutes -} The summarization process is a representation of the beliefs of the agent with regard to what is going on in its space over the course of an activity. Since the agent already needs to keep track of this information in order to make its decisions effectively, we can replay or sample from it to generate an automated visual summary of (the agent's belief of) the proceedings in the room.  
\item[-] {\em Decision Making Process -} Finally, and perhaps most importantly, the decision making process itself needs efficient interfacing with the humans -- this can involve a range of things from showing alternative solutions to a task, to justifying the reasoning behind different suggestions. 
This is crucial in a mixed initiative planning setting \cite{horvitz1999principles,horvitz2007reflections} to allow for human participation in the planning process, as well as for the planner's participation in the humans' decision making process.
\end{itemize}

\section{Mind of \jones}

\begin{figure}
\centering
\includegraphics[width=\columnwidth]{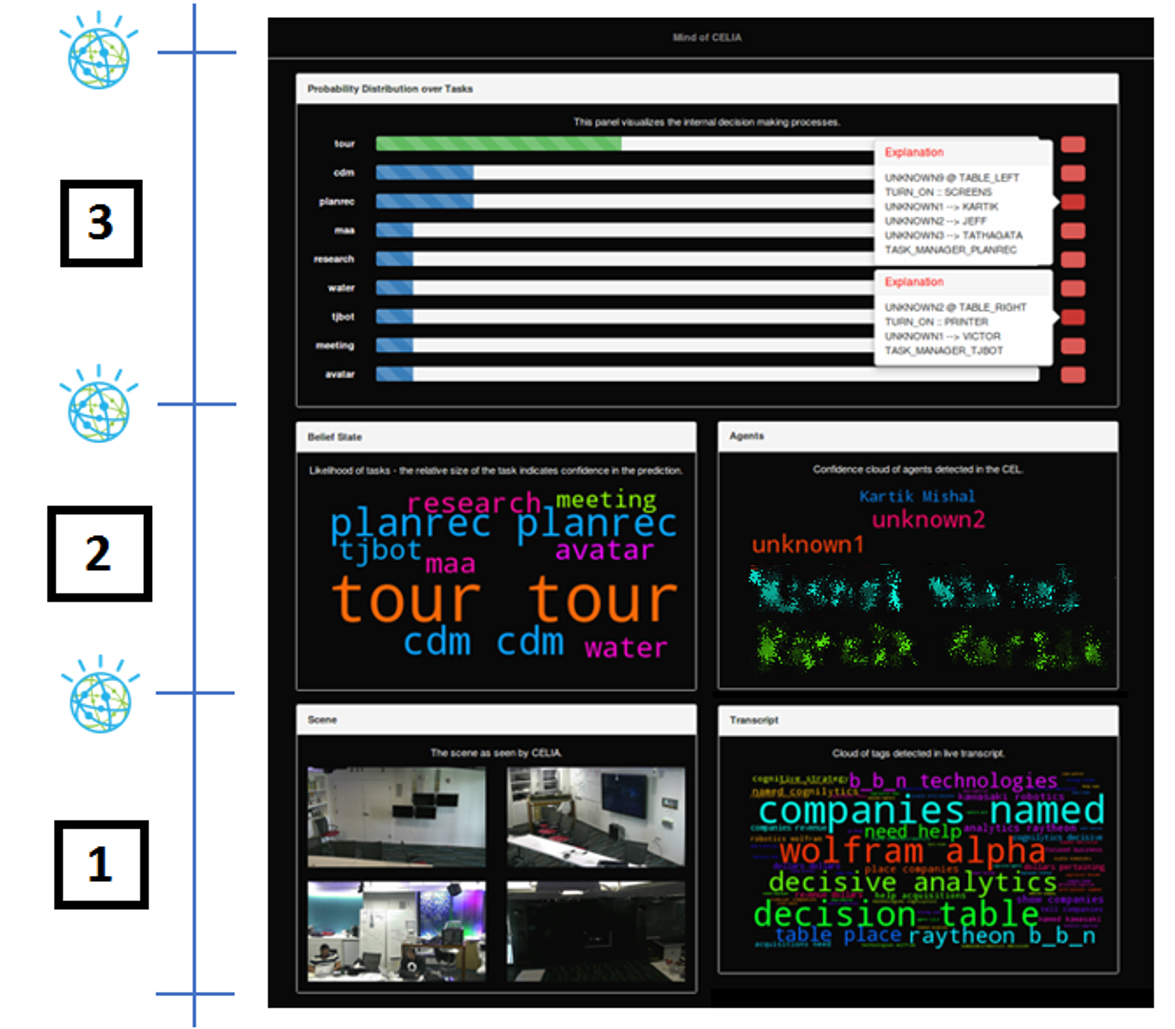}
\caption{Snapshot of the mind of \jones~externalizing different stages of its cognitive processes.}
\label{fig:snap}
\end{figure}

First, we will describe the externalization of the ``mind'' of \jones~-- i.e. the various processes that feed the different capabilities of the agent.
A snapshot of the interface is presented in Figure~\ref{fig:snap}. 
The interface itself consists of five widgets. The largest widget on the top shows the various usecases that the \cel~is currently set up to support. In the current \cel~setup, there are nine such usecases. The widget represents the probability distribution that indicates the confidence of \jones~in the respective task being the one currently being collaborated on, along with a button for the provenance of each such belief. 
%
%
The information used as provenance is generated directly from the plans used internally by the recognition module \cite{ramirez2010probabilistic}  and justifies why, given its model of the underlying planning problems, these tasks look likely in terms of plans that achieve those tasks.
%
%
%
Model based algorithms are especially useful in providing explanations like this \cite{sohrabi2011preferred,danmaga}. 
The system is adept at handling uncertainty in its inputs (it is interesting to note that in coming up with an explanatory plan it has announced likely assignments to unknown agents in its space). 
%
%
In Figure~\ref{fig:snap}, \jones~has placed the maximum confidence in the \texttt{tour} usecase.

Below the largest widget is a set of four widgets, each of which give users a peek into an internal component of \jones. The first widget, on the top left, presents a wordcloud representation of \jones's belief in each of the tasks; the size of the word representing that task corresponds to the probability associated with that task.  The second widget, on the top right, shows the agents that are recognized as being in the environment currently -- this information is used by the system to determine what kind of task is more likely. This information is obtained from four independent camera feeds that give \jones~an omnispective view of the environment; this information is represented via snapshots (sampled at 10-20 Hz) in the third widget, on the bottom left. In the current example, \jones~has recognized the agents named (anonymized) ``XXX'' and ``YYY'' in the scenario. Finally, the fourth widget, on the bottom right, represents a wordcloud based summarization of the audio transcript of the environment. This transcript provides a succinct representation of the things that have been said in the environment in the recent past via the audio channels. Note that this widget is merely a summarization of the full transcript, which is fed into the IBM Watson Conversation service to generate observations for the plan recognition module. The interface thus provides a (constantly updating) snapshot of the various sensory and cognitive organs associated with \jones~-- the eyes, ears, and mind of the \cel. 
This snapshot is also organized at increasing levels of abstraction -- 

\begin{itemize}
\item[{[1]}] {\em Raw Inputs --} These show the camera feeds and voice capture (speech to text outputs) as received by the system. These help in externalizing what information the system is working with at any point of time and can be used, for example, in debugging at the input level if the system makes a mistake or in determining whether it is receiving enough information to make the right decisions. It is especially useful for an agent like \jones, which is not embodied in a single robot or interface but is part of the environment as a whole. As a result of this, users may find it difficult to attribute specific events and outcomes to the agent. 
%
%
%
%
%
%
\item[{[2]}] {\em Lower level reasoning -- } 
The next layer deals with the first stage of reasoning over these raw inputs -- What are the topics being talked about? Who are the agents in the room? Where are they situated? This helps an user identify what knowledge is being extracted from the input layer and fed into the reasoning engines. It increases the situational awareness of agents by visually summarizing the contents of the scene at any point of time.
%
%
%
%
\item[{[3]}] {\em Higher level reasoning -- } 
Finally, the top layer uses information extracted at the lower levels to reason about abstract tasks in the scene. It visualizes the outcome of the plan recognition process, along with the provenance of the information extracted from the lower levels (agents in the scene, their positions, speech intents, etc.). This layer puts into context the agent's current understanding of the processes in the scene.
%
%
%
%
%
%
%
\end{itemize}

\begin{figure*}
\centering
\begin{subfigure}[b]{0.77\textwidth}
\includegraphics[width=\textwidth]{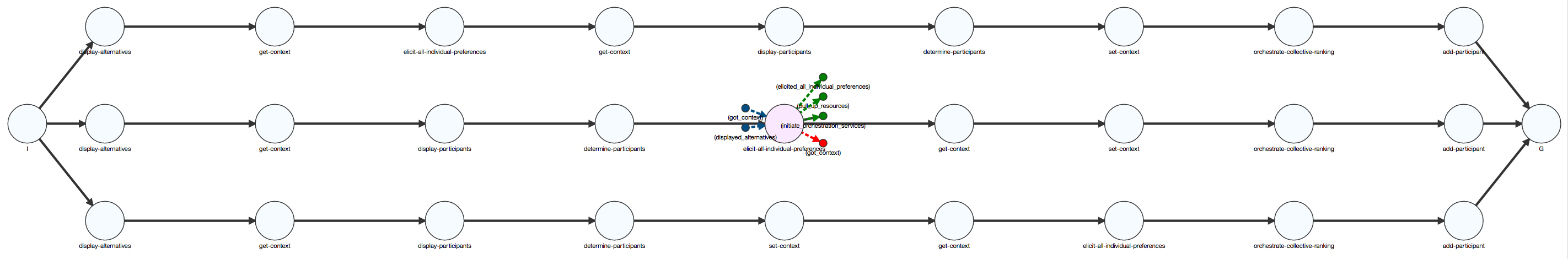}
\caption{Top-K plan visualization showing alternative plans for a given problem.}
\label{fig:subfig:1}
\end{subfigure}
\begin{subfigure}[b]{0.22\textwidth}
\includegraphics[width=\textwidth]{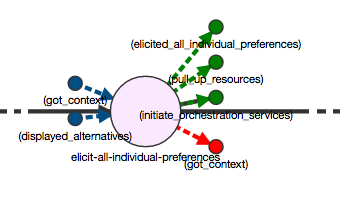}
\caption{Action Descriptions}
\label{fig:subfig:2}
\end{subfigure}
\caption{Visualization of plans in \fresco~showing top-K alternative solutions (K=3) for a given planing problem (left) and on-demand visualization of each action in the plan (zoomed-in; right) in terms of causal links consumed and produced by it.}
\label{fig:fig}
\end{figure*}

\begin{figure*}
\centering
\includegraphics[width=\textwidth]{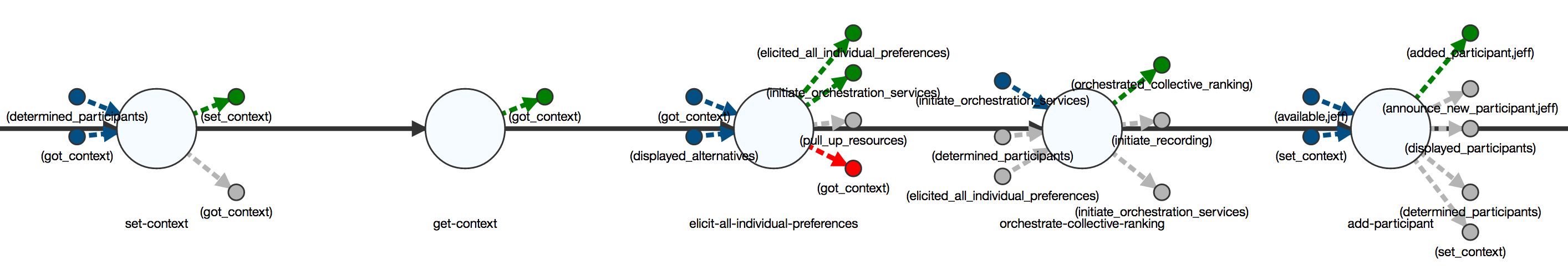}
\caption{Visualization as a process of explanation -- minimized view of conditions relevant to a plan. Blue, green and red nodes indicate preconditions, add and delete effects respectively. The conditions which are not necessary causes for this plan (i.e. the plan is still optimal in a domain without these conditions) are grayed out in the visualization (11 out of a total 30).}
\label{fig:fog}
\end{figure*}

\vspace{-10pt}
\paragraph{Demonstration 1}
We now demonstrate how the {\em Engage} process evolves as agents interact in the \cel. The demonstration begins with two humans discussing the \cel~ environment, followed by one agent describing a projection of the Mind of \jones~on the screen. The other agent then discusses how a Mergers and Acquisitions (\ma) task~\cite{kephart2015symbiotic} is carried out. 
%
%
{\em A video of this demonstration can be accessed 
at \protect\url{https://www.youtube.com/watch?v=ZEHxCKodEGs}.}
The video contains a window that demonstrates the evolution of the \jones~ interface through the duration of the interaction. This window illustrates how \jones's beliefs evolve dynamically in response to interactions in real-time. 

\vspace{-10pt}
\paragraph{Demonstration 2}
After a particular interaction is complete \jones\ can automatically compile a summarization (or minutes) of the meeting by sampling from the visualization of its beliefs. 
{\em An anonymized video of a typical summary can be accessed at \protect\url{https://youtu.be/AvNRgsvuVOo}}.
This kind of {\em visual summary} provides a powerful alternative to established meeting summarization tools like text-based minutes. The visual summary can also be used to extract abstract insights about this one meeting, or a set of similar meetings together and allows for agents that may have missed the meeting to catch up on the proceedings. Whilst merely sampling the visualization at discrete time-intervals serves as a powerful tool towards automated summary generation, we anticipate the use of more sophisticated visualization~\cite{5613451} and summarization~\cite{wimbledon,kim2015inferring-jair,kim2016improving} techniques in the future.

\section{Model-Based Plan Visualization : \fresco}

We start by describing the planning domain that is used in the rest of this section, followed by a description of \fresco's different capabilities in terms of {\em top-K plan visualization} and {\em model-based plan visualization}. We conclude by describing the implementation details on the back-end.

\vspace{-10pt}
\paragraph{The Collective Decision Domain}
We use a variant of the Mergers and Acquisitions (\ma) task called \emph{Collective Decision} (CD). 
The CD domain models the process of gathering input from a decision makers in a smart room, and the orchestration of comparing alternatives, eliciting preferences, and finally ranking of the possible options. 

\subsection{Top-K Visualization}

Most of the automated planning technology and literature considers the problem of generating a single plan. Recently, however, the paradigm of Top-K planning~\cite{riabov2014new} has gained traction. Top-K plans are particularly useful in domains where producing and deliberating on multiple alternative plans that go from the same fixed initial state and the same fixed goal is important. Many decision support scenarios, including the one described above, are of this nature. Moreover, Top-K plans can also help in realizing unspecified user preferences, which may be very hard to model explicitly. By presenting the user(s) with multiple alternatives, an \emph{implicit} preference elicitation can instead be performed. The \fresco\ interface supports visualization of the $K$ top plans for a given problem instance and domain model, as shown in Figure~\ref{fig:subfig:1}. In order to generate the Top-K plans, we use an experimental Top-K planner~\cite{katz2017blind} that is built on top of Fast Downward~\cite{helmert2006fast}.


\subsection{Model-based Plan Visualization}

The requirements for visualization of plans can have different semantics depending on the task at hand -- e.g. showing the search process that produced the plan, and the decisions taken (among possible alternative solutions) and trade-offs made (by the underlying heuristics) in that process; or revealing the underlying domain or knowledge base that engendered the plan. The former involves visualizing the {\em how} of plan synthesis,  while the latter focuses on the {\em why}, and is model-based and algorithm independent. Visualizing the how is useful to the developer of the system during debugging, but serves little purpose for the end user who would rather be told the rationale behind the plan: why is this plan better than others, what individual actions contribute to the plan, what information is getting consumed at each step, and so on. Unfortunately, much of the visualization work in the planning community has been confined to depicting the search process alone~\cite{thayerYoutube,thayer2012heuristic,magnaguagno2017webplanner}. 
\fresco, on the other hand, aims to focus on the \emph{why} of a plan's genesis, in the interests of establishing common ground with human decision-makers. 
At first glance, this might seems like an easy problem -- we could just show what the preconditions and effects are for each action along with the causal links in the plan. However, even for moderately sized domains, this turns into a clumsy and cluttered approach very soon, given the large number of conditions to be displayed. 
In the following, we will describe how \fresco\ handles this problem of overload.

\vspace{-10pt}
\paragraph{Visualization as a Process of Explanation}
We begin by noting that the process of visualization can in fact be seen as a {\em process of explanation}. 
In model-based visualization, as described above, the system is essentially trying to explain to the viewer the salient parts of its knowledge that contributed to this plan. In doing so, it is externalizing what each action is contributing to the plan, as well as outlining why this action is better that other possible alternatives. 

\vspace{-10pt}
\paragraph{Explanations in Multi-Model Planning}
Recent work has shown~\cite{explain} how an agent can explain its plans to the user when there are differences in the models (of the same planning problem) of the planner and the user, which may render an optimal plan in the planner's model sub-optimal or even invalid--and hence unexplainable--in the user's mental model. 
%
%
%
%
An explanation in this setting constitutes a {\em model update} to the human such that the plan (that is optimal to the planner) in question also becomes optimal in the user's updated mental model. This is referred to as a {\em model reconciliation process} (MRP). The smallest such explanation is called a {\em minimally complete explanation} (MCE).

\begin{figure*}
\centering
\begin{subfigure}[b]{0.75\textwidth}
\centering
\includegraphics[width=\textwidth]{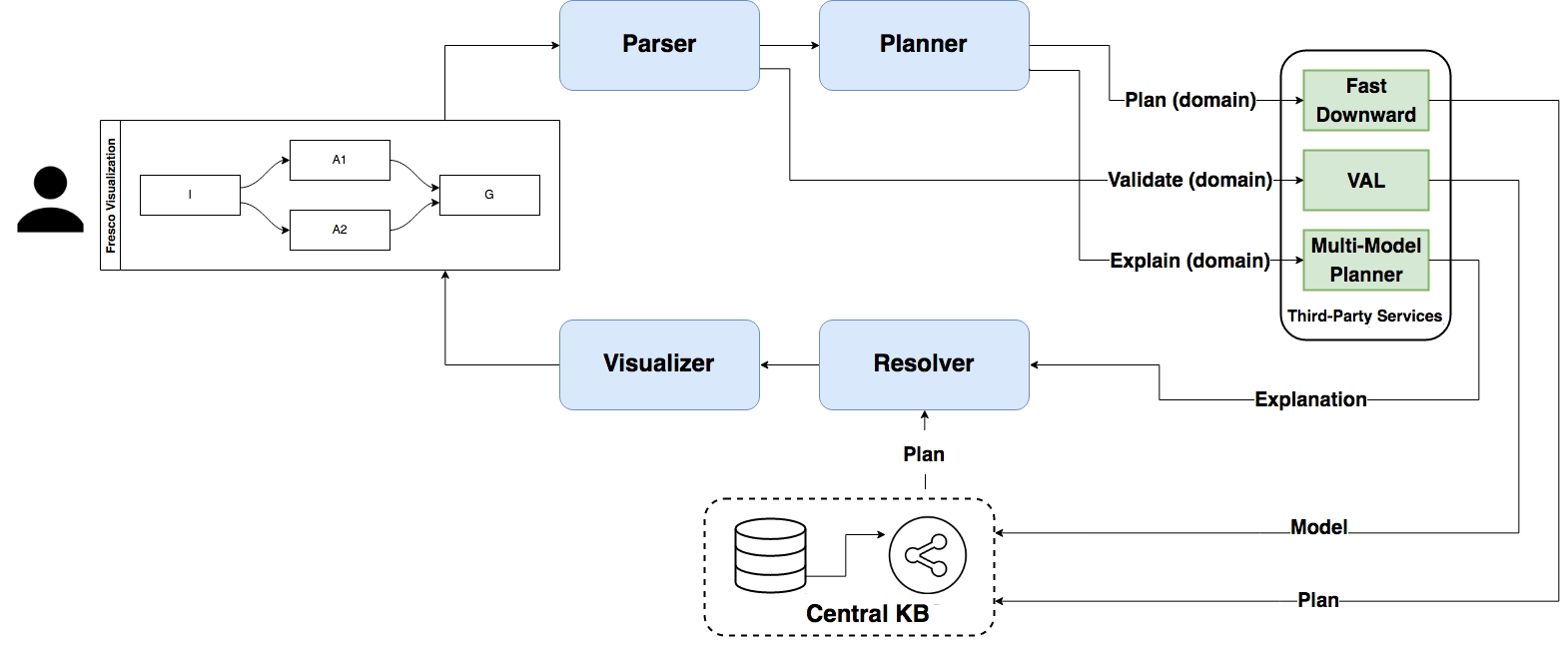}
\caption{Architecture diagram of \fresco.}
\label{fig:subfig:2}
\end{subfigure}
\qquad
\begin{subfigure}[b]{0.2\textwidth}
\centering
\includegraphics[width=\textwidth]{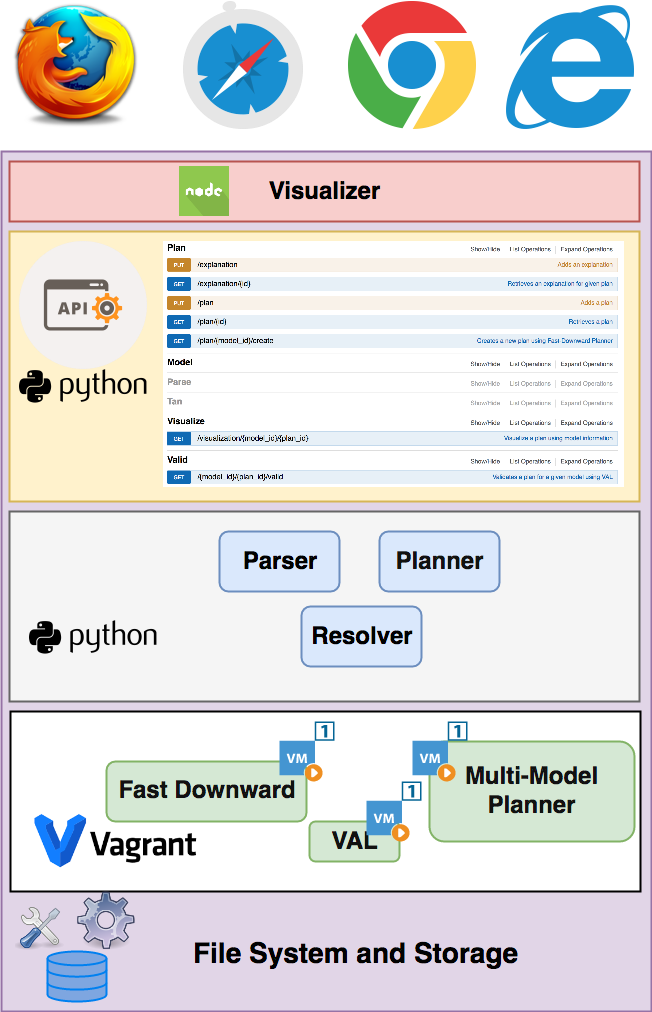}
\caption{Software stack}
\label{fig:subfig:3}
\end{subfigure}
\caption{Illustration of the flow of control (left) in \fresco~ between the plan generator (FD), explanation generator (MMP), and plan validator (VAL) with the visualization modules. The MMP code base is in the process of being fully integrated into \fresco, and it is currently run as a stand-alone component. The software stack (right) shows the infrastructure supporting \fresco~ in the backend.}
\label{fig:fig2}
\end{figure*}

\vspace{-10pt}
\paragraph{Model-based Plan Visualization $\equiv$ Model Reconciliation with Empty Model}

As we mentioned previously, exposing the entire model to the user is likely to lead to cognitive overload and lack of situational awareness due to the amount of information that is not relevant to the plan in question. 
We want to minimize the clutter in the visualization and yet maintain all relevant information pertaining to the plan.
{\em We do this by launching an instantiation of the model reconciliation process with the planner's model and an empty model as inputs.}
An empty model is a copy of the given model where actions do not have any conditions and the initial state is empty (the goal is still preserved).
Following from the above discussion, the output of this process is then the minimal set of conditions in the original model that ensure optimality of the given plan. 
In the visualization, the rest of the conditions from the domain are grayed out. 
\cite{explain} showed how this can lead to a significant pruning of conditions that do not contribute to the generation of a particular plan.
An instance of this process on the CD domain is illustrated in Figure~\ref{fig:fog}. 

Note that the above may not be the only way to minimize information being displayed. 
There might be different kinds of information that the user cares about, depending on their preferences. This is also highlighted by the fact that an MCE is not unique for a given problem. These preferences can be {\em learned} in the course of interactions.


\subsubsection{Architecture of \fresco}

The architecture of \fresco, shown in Figure~\ref{fig:subfig:2},  includes several core modules such as the parser, planner, resolver, and visualizer. These modules are all connected in a feed-forward fashion. 
The parser module is responsible for converting domain models and problem instances into python objects, and for validating them using VAL~\cite{howey2004val}. Those objects are then passed on to the planner module, which relies on Fast-Downward (FD) and the Multi-Model Planner (MMP)~\cite{explain} to generate a plan along with its explanation. 
The resolver module consumes the plan, the explanation, and the domain information to not only ground the plan, but also to remove any preconditions, add, or delete effects that are deemed irrelevant by the MMP module. 
Finally, the visualizer module takes the plan from the resolver module as an input, and builds graphics that can be rendered within any well-known web browser. Our focus in designing the architecture was on making it functionally modular and configurable, as shown in Figure~\ref{fig:subfig:3}. While the first three modules described above are implemented using Python, the visualizer module is implemented using Javascript and the \texttt{D3} graphics library. Our application stack uses \texttt{REST} protocols to communicate between the visualizer module and the rest of the architecture. We also accounted for scalability and reliability concerns by containerizing the  application with \texttt{Kubernetes}, in addition to building individual containers / virtual machines for third party services like VAL, Fast-Downward, and MMP.  

\newpage

\section{Work in Progress} 

While we presented the novel notion of {\em explanation as visualization} in the context of AI planning systems in this paper via the implemention of the \jones~assistant, there is much work yet to be done to embed this as a central research topic in the community. 
We conclude the paper with a brief outline of future work as it relates to the visualization capabilities of \jones~and other systems like it.

\vspace{-10pt}
\paragraph{Visualization for Model Acquisition}
Model acquisition is arguably the biggest bottleneck in the widespread adoption of automated planning technologies. Our own work with \jones\ is not immune to this problem. Although we have enabled an XML-based modeling interface, the next iteration of making this easily consumable for non-experts involves two steps: first, we impose an (possibly graphical) interface on top of the XML structure to obtain information in a structured manner. We can thenl provide visualizations such as those described in~\cite{bryce2017insitu} in order to help with iterative acquisition and refinement of the planning model.

\vspace{-10pt}
\paragraph{Tooling Integration}
Eventually, our vision -- not restricted to any one planning tool or technology -- is to integrate the capabilities of \fresco\ into a domain-independent planning tool such as \texttt{planning.domains}~\cite{muise-icaps16demo-pd}, which will enable the use of these visualization components across various application domains.
\texttt{planning.domains} realizes the long-awaited planner-as-a-service paradigm for end users, but is yet to incorporate any visualization techniques for the user. 
Model-based visualization from \fresco, complemented with search visualizations from emerging techniques like \texttt{WebPlanner}~\cite{magnaguagno2017webplanner}, can be a powerful addition to the service.

\vspace{-10pt}
\paragraph{Acknowledgements}
A significant part of this work was initiated
and completed while Tathagata Chakraborti was an
intern at IBM’s T. J. Watson Research Center during 
the summer of 2017. The continuation
of his work at ASU is supported by an IBM Ph.D.
Fellowship.

\cleardoublepage
\bibliographystyle{named}
\bibliography{ijcai18}

\end{document}